\documentclass[letterpaper]{article} 
\usepackage{aaai2027}  
\usepackage[hyphens]{url}  
\usepackage{graphicx} 
\usepackage{natbib}  
\usepackage{caption} 
\usepackage{algorithm}
\usepackage{algorithmic}

\usepackage{newfloat}
\usepackage{listings}
\DeclareCaptionStyle{ruled}{labelfont=normalfont,labelsep=colon,strut=off} 
\floatstyle{ruled}
\newfloat{listing}{tb}{lst}{}
\floatname{listing}{Listing}

\usepackage{booktabs}
\usepackage{multirow}

\title{EvoMem: Self-Evolving Skills for Time-Aware Long Video Memory Agents}
\author{
    Yutong Guan\textsuperscript{\rm 1},
    Meixing Shi\textsuperscript{\rm 1},
    Haoran Sun\textsuperscript{\rm 2},
    Bing Sun\textsuperscript{\rm 3},
    Yuxiang Cai\textsuperscript{\rm 1},
    Yankai Jiang\textsuperscript{\rm 2}
}
\affiliations{
    \textsuperscript{\rm 1}Zhejiang University\\
    \textsuperscript{\rm 2}Shanghai Artiffcial Intelligence Laboratory \\
    \textsuperscript{\rm 3}Research Center, National Certiffcation Technology (Hangzhou) Co., Ltd\\
   
}

\begin{document}

\maketitle

\begin{abstract}
Although memory-augmented architectures have expanded the ability of large vision-language models (LVLMs) to process long videos, existing retrieval processes still largely rely on global semantic matching and lack state-driven temporal control, often returning fragmented or redundant evidence and failing to maintain a coherent evidence chain. To address this challenge, we propose \textbf{EvoMem}, a time-aware long-video memory agent guided by self-evolving retrieval skills. EvoMem constructs a hierarchical time-indexed memory, where different types of evidence are aligned onto a unified video timeline; during retrieval, temporal cues from acquired evidence are further converted into explicit signals that constrain subsequent memory access. To address recurring procedural errors in long-video question answering, such as unclear retrieval targets, improper memory access paths, or premature stopping, EvoMem introduces retrieval skills to provide process-level guidance, enabling the agent to dynamically adjust retrieval content, memory operations, and target temporal ranges according to the current evidence state, while deciding when to stop. Furthermore, inspired by reinforcement learning, we formulate skill evolution as a reward-guided policy improvement process. Each candidate skill is treated as an external policy patch, and rewards are assigned according to answer-correctness state transitions on the validation set. In this way, effective skills are accepted and low-quality skills are rejected without updating model parameters. Experiments on four long-video question answering benchmarks show that EvoMem significantly outperforms existing baselines, achieving an average accuracy of 60.1\% and an average improvement of 8.6 percentage points over the previous best methods, demonstrating its effectiveness for long-video reasoning.
\end{abstract}


\section{Introduction}
\begin{figure*}[t]
  \centering
  \includegraphics[width=\textwidth]{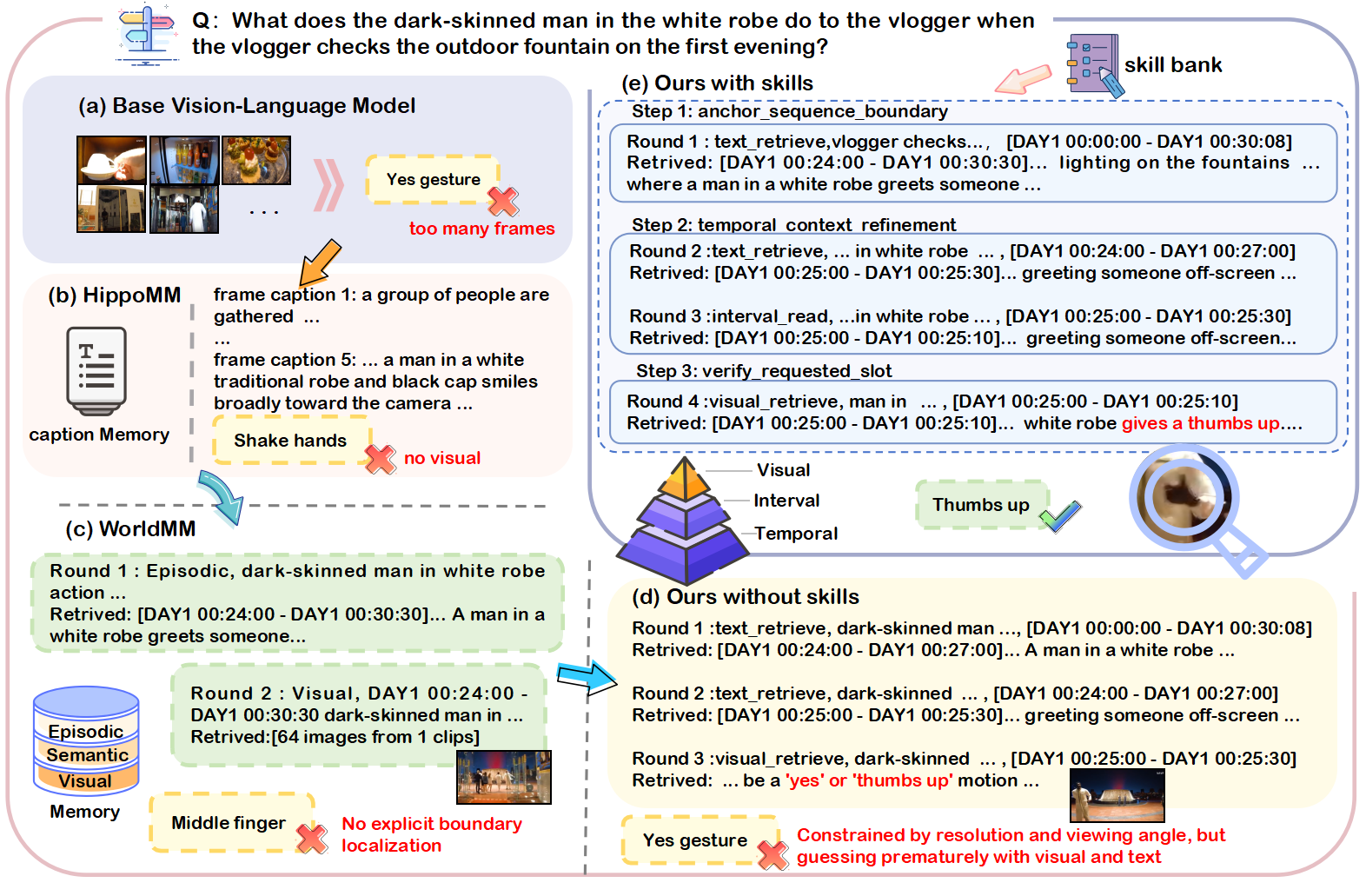}
  \caption{Motivating example of EvoMem. (a) Base VLMs suffer from redundant frames and limited context. (b) HippoMM's text-centric semantic memory may miss fine-grained visual evidence. (c) WorldMM lacks explicit temporal control for maintaining evidence chains. (d) Skill-free retrieval lacks guidance on what to seek, where to search, and when to stop. (e) EvoMem uses self-evolving skills for time-aware memory retrieval and coherent long-video reasoning.}
  \label{fig:motivation}
\end{figure*}

Recent advances in large vision-language models (LVLMs) have substantially improved video understanding, particularly on short clips with limited temporal scope~\cite{bai2025qwen3vltechnicalreport,comanici2025gemini25pushingfrontier,wang2025internvl35advancingopensourcemultimodal,lin2024video,li2024llava}. However, extending these capabilities to hour-scale or longer videos~\cite{yang2025egolife,tian2025egor1chainoftoolthoughtultralongegocentric,Wang_2025_ICCV,Fu_2025_CVPR,NEURIPS2024_329ad516} remains fundamentally challenging: question-relevant evidence is often sparse, temporally scattered, and buried in redundant visual and textual observations, while correct answering requires modeling temporal dependencies across distant events. Simply scaling up the context window with more frames, captions, or intermediate summaries~\cite{liu2025videoxlproreconstructivetokencompression,li2025videochatflashhierarchicalcompressionlongcontext,wang2026lvc} introduces high computational cost and severe distraction from irrelevant context, as illustrated in Fig.~\ref{fig:motivation}(a). As a result, even strong LVLMs often struggle to maintain stable temporal grounding and precise evidence localization in long videos.

To overcome the limitations of direct long-context modeling, recent studies have increasingly shifted toward memory-augmented architectures and retrieval-based video agents. Video-RAG organizes long-video content into queryable evidence units through retrieval augmentation~\cite{NEURIPS2025_f593c9c2,wang2024videoagent}. HippoMM builds hippocampal-inspired multimodal memories by consolidating audiovisual streams into semantic representations, while WorldMM constructs complementary episodic, semantic, and visual memories for multimodal retrieval across temporal scales~\cite{lin2026hippomm, yeo2026worldmm}. In parallel, agentic systems such as M3Agent and Ego-R1 introduce tool use and multi-step reasoning, enabling LVLMs to iteratively acquire question-relevant information~\cite{long2025seeinglisteningrememberingreasoning, tian2025egor1chainoftoolthoughtultralongegocentric}. Together, these efforts move long-video understanding toward selective access to external memories, laying an important foundation for scalable long-video reasoning systems.

However, as illustrated in Fig.~\ref{fig:motivation}(b) and Fig.~\ref{fig:motivation}(c), the availability of external memory does not necessarily lead to reliable evidence retrieval. Agent systems that rely on textual summaries or structured nodes can improve evidence accessibility, but they also compress low-level visual information, making it difficult to verify fine-grained motions, object states, or local visual differences. Visual memory access can complement these visual cues to some extent; however, as shown in Fig.~\ref{fig:motivation}(c), when visual retrieval still relies on a coarse temporal range obtained from global textual matching, visual verification may be performed over an imprecise candidate window and miss the key evidence. These issues suggest that when upstream textual or semantic retrieval still operates over the global memory space, multi-turn retrieval can be attracted to semantically similar segments and repeatedly return temporally fragmented or redundant evidence, thereby failing to continuously complete the evidence chain within the target temporal range. Therefore, a key bottleneck of existing methods lies not only in memory construction or tool availability, but also in the lack of a state-driven temporal retrieval mechanism that can convert accumulated evidence into temporal constraints for subsequent memory access.

To address the above limitations, we propose \textbf{EvoMem}, a time-aware long-video memory agent that combines state-driven temporal retrieval with a self-evolving retrieval skill bank, as shown in Fig.~\ref{fig:motivation}(d) and Fig.~\ref{fig:motivation}(e). EvoMem first organizes long videos into hierarchical time-indexed memories, enabling the agent to convert temporal cues obtained during retrieval into explicit constraints and dynamically narrow, expand, or shift subsequent retrieval ranges, thereby continuously acquiring evidence along the same temporal evidence chain. However, long-video questions often involve complex and diverse evidence requirements, and different question types correspond to different retrieval targets, modality preferences, retrieval orders, and stopping conditions. To prevent multi-turn retrieval from degenerating into locally guided trial-and-error without global planning, and to equip the agent with cross-turn retrieval planning ability, EvoMem selects a relevant retrieval skill before inference begins to guide target localization, memory operations, temporal range adjustment, semantic queries, and stopping decisions during multi-step retrieval. Furthermore, EvoMem formulates skill self-evolution as reward-guided policy improvement: candidate skills are induced from successful and failed reasoning trajectories and evaluated on independent validation questions against the current active skill bank; skills that repair previous errors receive positive rewards, while those that cause performance regressions are penalized. Only candidate skills with sufficiently high cumulative validation rewards are accepted. In this way, EvoMem continuously optimizes the external retrieval policy without updating foundation model parameters, turning accepted skills into reusable procedural experience for subsequent long-video reasoning.

The main contributions of this work are as follows:

\begin{itemize}
    \item We propose \textbf{EvoMem}, a time-aware long-video memory agent guided by self-evolving retrieval skills. EvoMem organizes long videos into hierarchical time-indexed video memories and uses controllable memory operations to turn temporal information into explicit constraints for subsequent memory access, enabling state-driven temporal evidence acquisition.
    
    \item We introduce a reinforcement-learning-inspired skill self-evolution mechanism, where candidate retrieval skills are treated as candidate updates to the current retrieval policy and are rewarded according to answer-correctness transitions during validation execution. This enables EvoMem to selectively accept effective skills, reject regressive skills, and continuously optimize the skill bank without updating the parameters of the foundation model.
    
    \item We systematically evaluate EvoMem on four long-video question answering benchmarks, and validate the effectiveness of time-aware retrieval, retrieval skill guidance, and reward-guided skill self-evolution through component ablations and cross-model skill generalization experiments.
\end{itemize}

\section{Related Work}

\subsection{Long Video Understanding}
Recent large vision-language models (LVLMs) have achieved substantial progress in video understanding and have gradually been extended from short clips to hour-scale or even longer videos~\cite{bai2025qwen3vltechnicalreport, comanici2025gemini25pushingfrontier, wang2025internvl35advancingopensourcemultimodal}. To mitigate the context pressure introduced by ultra-long videos, existing studies mainly improve long-context visual modeling~\cite{bai2025qwen3vltechnicalreport, comanici2025gemini25pushingfrontier, zhang2024longcontexttransferlanguage} or reduce redundant video content through spatiotemporal sampling, hierarchical summarization, and token compression~\cite{liu2025videoxlproreconstructivetokencompression, yang2025pvc,wang2026lvc}. However, they still largely rely on preserving or compressing video information within a limited context, making it difficult to avoid sparse key evidence being overwhelmed by redundant content. This motivates a growing shift toward selectively acquiring and using question-relevant evidence in long-video reasoning.

\subsection{Memory-Augmented Video LLMs}

To support long-video reasoning, a growing body of work augments video language models with external retrieval and memory mechanisms. Early retrieval-augmented methods typically convert long videos into queryable frames, clips, textual descriptions, or graph nodes, and retrieve question-relevant evidence during inference rather than relying solely on uniformly sampled video inputs. Building on this direction, recent methods have evolved from retrieval-augmented evidence acquisition~\cite{NEURIPS2025_f593c9c2,ma2025drvideo,NEURIPS2025_092359ce} to tool-using agentic reasoning~\cite{tian2025egor1chainoftoolthoughtultralongegocentric} and explicit multimodal video memory construction~\cite{lin2026hippomm, long2025seeinglisteningrememberingreasoning,yeo2026worldmm}, collectively improving the accessibility of long-video evidence. However, these methods mainly focus on constructing and accessing external evidence, while lacking mechanisms to actively adjust temporal ranges, memory sources, and retrieval processes according to the evolving evidence state.

\subsection{Skill-Based LLM Agents}

Recent studies on LLM agents have begun to treat skills as reusable, manageable, and continuously updatable procedural knowledge~\cite{yang2026autoskillexperiencedrivenlifelonglearning,wang2026skillxautomaticallyconstructingskill,xia2026skillrlevolvingagentsrecursive,ma2026skillclawletskillsevolve,zhou2026mementoskillsletagentsdesign}. MUSE-Autoskill organizes skill creation, memory, management, evaluation, and refinement into a unified lifecycle~\cite{lin2026museautoskillselfevolvingagentsskill}; SkillOps studies defect diagnosis, compatibility, and long-term usability of skill libraries from a library-maintenance perspective~\cite{pu2026skillopsmanagingllmagent}. Relatedly, MemSkill represents agent memory operations as learnable and evolvable memory skills~\cite{zhang2026memskilllearningevolvingmemory}. Unlike these works, which mainly focus on general task execution, skill-library maintenance, or general memory management, EvoMem studies self-evolving skills for long-video temporal memory retrieval and continuously improves the skill bank through reinforcement-learning-inspired validation feedback.

\section{Method}

\begin{figure*}[t]
  \centering
  \includegraphics[width=\textwidth]{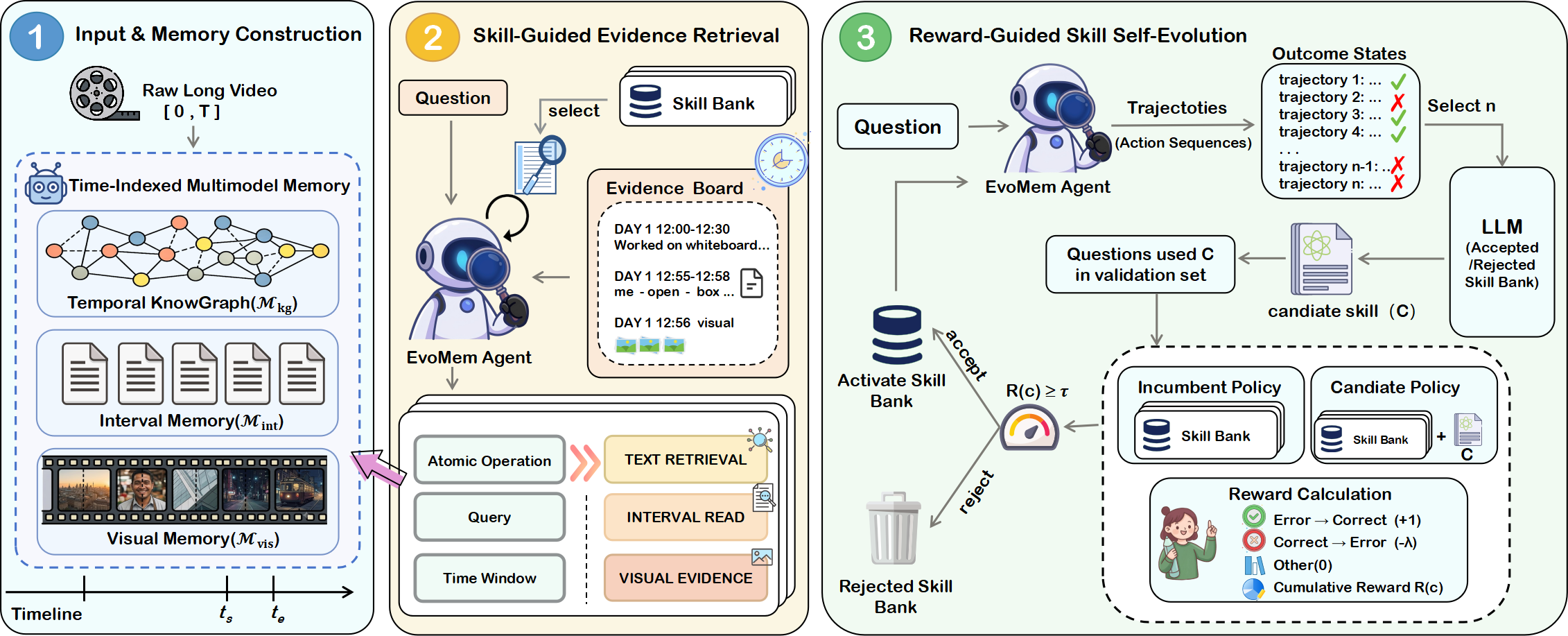}
  \caption{Overall pipeline of EvoMem: hierarchical time-indexed memory construction, skill-guided retrieval over textual, interval, and visual evidence, and reward-guided active skill bank updates without modifying model parameters.}
  \label{fig:Overall}
\end{figure*}

This section presents the overall framework of EvoMem, as shown in Fig.~\ref{fig:Overall}. Given a long video and a question, EvoMem first constructs hierarchical time-indexed memories, enabling structured relations, interval textual context, and visual evidence to be accessed on a unified video timeline. It then performs skill-guided state-adaptive evidence retrieval, where the agent dynamically adjusts retrieval content, memory operations, and target temporal ranges according to the current evidence state. Finally, EvoMem uses reward-guided skill self-evolution to transform procedural feedback from successful and failed trajectories into reusable retrieval skills, thereby continuously optimizing the external retrieval policy.

\subsection{Time-Indexed Video Memory Construction}

In long videos, similar events, human actions, and object states may recur at different temporal positions, making retrieval based on semantic matching alone prone to conflating temporally distant segments. To address this, EvoMem associates event relations, interval summaries, and visual evidence with explicit temporal boundaries, making time a key basis for distinguishing recurring events and organizing evidence chains. Based on this, EvoMem converts the input video $\mathcal{V}$ into a hierarchical time-indexed video memory bank $\mathcal{M}$, providing a unified basis for subsequent time-aware memory access and cross-modal evidence alignment. Formally, the complete memory bank consists of three complementary memories:
\begin{equation}
\mathcal{M}=\mathcal{M}_{kg}\cup\mathcal{M}_{int}\cup\mathcal{M}_{vis}.
\end{equation}

First, EvoMem constructs the structured textual memory as a multi-scale temporal knowledge graph $\mathcal{M}_{kg}$. Continuous textual summaries preserve contextual information, but in long videos, relevant events are often buried in large amounts of narrative detail, and similar actions or repeated events may appear at distant temporal positions. To address this, EvoMem generates video summaries at multiple temporal scales $\mathcal{L}=\{\ell_1,\ldots,\ell_K\}$ and extracts time-bounded event triplets from the summaries at each scale. These triplets represent people, actions, objects, locations, and state changes as structured graph edges with temporal boundaries. The memory is defined as:
\begin{equation}
\mathcal{M}_{\mathrm{kg}}
=
\left\{
m_{i,\mathrm{kg}}
=
\bigl(h_i, r_i, o_i, [t_i^s,t_i^e], \ell_i\bigr)
\right\}_{i=1}^{N_{\mathrm{kg}}}.
\end{equation}
where $h_i$, $r_i$, and $o_i$ denote the head entity, relation, and tail entity, respectively. The interval $[t_i^s,t_i^e]$ is the source video span of the graph edge, and $\ell_i\in\mathcal{L}$ indicates its temporal scale. Fine-scale edges preserve local actions, object states, and interaction details, while coarse-scale edges summarize the major activity structure over longer temporal ranges. Through this graph memory, event cues scattered across different segments can be explicitly connected through entities, actions, and relations, while similar events remain distinguishable through their temporal boundaries.

Second, EvoMem maintains a finest-scale interval memory $\mathcal{M}_{int}$ for interval reading. Unlike the temporal knowledge graph, interval memory directly stores continuous textual summaries at the finest temporal scale:
\begin{equation}
\mathcal{M}_{\mathrm{int}}
=
\left\{
m_{j,\mathrm{int}}
=
\bigl(c_j, [t_j^s,t_j^e], \ell_{\min}\bigr)
\right\}_{j=1}^{N_{\mathrm{int}}}.
\end{equation}
where $c_j$ denotes the textual description of a basic temporal segment. All interval memory units are ordered along the original video timeline, preserving contextual continuity between adjacent segments. Once a candidate temporal window has been located, this memory allows the system to recover fine-grained local event progression, temporal order, and option-related details from the corresponding summary sequence.

Finally, we construct a time-aligned visual memory $\mathcal{M}_{\mathrm{vis}}$ to preserve visual details that cannot be fully captured by textual summaries or triplet extraction. Specifically, we divide the long video into short video segments and use a multimodal encoder to extract a visual feature for each segment. Each visual memory unit stores the segment-level visual feature together with its temporal boundary:
\begin{equation}
\mathcal{M}_{\mathrm{vis}}
=
\left\{
m_{k,\mathrm{vis}}
=
\bigl(\mathbf{v}_k,[t_k^s,t_k^e]\bigr)
\right\}_{k=1}^{N_{\mathrm{vis}}}.
\end{equation}
The term $\mathbf{v}_k$ denotes the visual feature of the $k$-th video segment, and $[t_k^s,t_k^e]$ denotes its temporal boundary in the original video. Visual memory preserves fine-grained visual cues such as object appearance, spatial relations, hand actions, object locations, and visible states, complementing the textual memories.

Together, these memories form a temporally addressable evidence space for subsequent time-aware retrieval and cross-modal evidence alignment.

\subsection{Skill-Guided State-Adaptive Evidence Retrieval}

After constructing the time-indexed video memory bank $\mathcal{M}$, EvoMem defines a set of atomic memory access operations with controllable temporal ranges over the memory bank. Specifically, the system supports three types of memory operations: textual retrieval over the multi-scale temporal knowledge graph $\mathcal{M}_{\mathrm{kg}}$ for discovering candidate events, entity relations, and state changes; interval reading over the interval textual memory $\mathcal{M}_{\mathrm{int}}$ for reading continuous local context within a target temporal range; and visual evidence acquisition over the visual memory $\mathcal{M}_{\mathrm{vis}}$ for verifying fine-grained visual cues that may be compressed or omitted by textual memories. These operations constitute the executable memory access space of EvoMem, enabling the agent to acquire evidence from different memory sources and temporal ranges.

Based on these atomic operations, the executor performs long-video question answering as a multi-turn evidence acquisition process. At retrieval turn $t$, the current state is represented as
\begin{equation}
s_t=(q,O,E_{t-1},H_{t-1}),
\end{equation}
where $q$ denotes the input question, $O$ denotes the candidate options, $E_{t-1}$ denotes the accumulated evidence state up to the previous turn, and $H_{t-1}$ denotes the retrieval history. The executor generates a retrieval decision based on the current state:
\begin{equation}
d_t=F_{\mathrm{exec}}(s_t).
\end{equation}
For a non-stop decision, $d_t$ is represented as $(a_t,u_t,w_t)$, corresponding to an atomic memory operation, a semantic query, and a target temporal window, respectively. If the evidence is sufficient, $d_t$ corresponds to a stop action, and the accumulated evidence is passed to the final answering module.

After executing a non-stop decision, the system retrieves the turn-level evidence from the corresponding memory:
\begin{equation}
\Delta E_t=a_t(u_t,w_t;\mathcal{M}).
\end{equation}
The retrieved evidence $\Delta E_t$ is then incorporated into the evidence state $E_t$, while the current action, query, temporal window, and retrieved evidence are appended to the retrieval history $H_t$. The updated state is then carried to the next retrieval turn.

On top of this base execution process, EvoMem introduces question-type-aware retrieval skills to provide process-level guidance for multi-turn evidence acquisition. For each question, the system selects a skill $S$ that matches the question description and candidate options, and keeps it fixed throughout the retrieval process. Each skill contains applicability conditions, exclusion conditions, staged process steps, and stopping conditions, and generates a process control signal from the current state. Formally, the decision of the skill-guided executor is written as:
\begin{equation}
d_t=F_{\mathrm{exec}}(s_t,S(s_t)).
\end{equation}
Here, $S(s_t)$ summarizes the current retrieval stage, the completeness of the evidence chain, evidence gaps to be filled, the next retrieval direction, and the stop decision. By incorporating this control signal into decision making, EvoMem can more effectively schedule textual retrieval, interval reading, and visual evidence acquisition along the same temporal evidence chain.

\subsection{Reward-Guided Skill Self-Evolution}

In complex long-video reasoning, repeated agent failures are often not incidental execution errors, but symptoms of systematic deficiencies in the underlying retrieval policy. For example, the agent may stop before the evidence chain is complete, fall into local or repetitive retrieval, fail to dynamically adjust the temporal range according to accumulated evidence, or overlook necessary visual verification when textual cues are ambiguous. To systematically address such procedural errors, EvoMem draws on the core reinforcement-learning idea of improving a policy through reward feedback, and formulates skill self-evolution as a reward-guided policy improvement process in the discrete space of retrieval skills.

The key idea is to lift retrieval skills from static prompts to an external policy state that can be continuously optimized. As described above, a selected retrieval skill $S$ generates a process control signal from the current state, $g_t=S(s_t)$, which further constrains the executor's decision $d_t=F_{\mathrm{exec}}(s_t,g_t)$. Therefore, reshaping the set of available skills changes the retrieval behavior of the agent under the same evidence state. Formally, EvoMem defines the set of accepted skills at evolution round $r$ as the active skill bank $\mathcal{B}^{(r)}$. The retrieval policy induced by this bank is denoted as the incumbent policy $\pi_{\mathcal{B}^{(r)}}$, which serves as a dynamic baseline for evaluating candidate skills. At the initial round, $\mathcal{B}^{(0)}$ is initialized as an empty accepted-skill bank; as evolution proceeds, only candidate skills that pass validation are added to this bank.

At evolution round $r$, EvoMem first deploys the incumbent policy $\pi_{\mathcal{B}^{(r)}}$ on a set of training tasks, collecting complete multi-turn retrieval trajectories, state transition records, and final answer correctness. These trajectories constitute the agent's experience from interacting with the long-video memory environment, and expose scheduling blind spots of the current policy. Based on this trajectory-level feedback, EvoMem induces a candidate skill $c$ to encode a reusable procedural correction. The candidate skill is not a memorized fix for a single failed example; instead, it generalizes the applicability boundary, staged progression logic, and stopping condition of a class of retrieval procedures. From the policy-evolution perspective, the candidate skill $c$ acts as a textual policy patch. Temporarily attaching it to the current active skill bank yields an exploratory candidate policy $\pi_{\mathcal{B}^{(r)}\cup\{c\}}$.

\begin{table*}[t]
\centering
\begin{tabular}{llcccccc}
\toprule
\textbf{Category} & \textbf{Method} & \textbf{Frames} & \textbf{EgoLifeQA} & \textbf{Ego-R1 Bench} & \textbf{LVBench} & \textbf{Video-MME (L)} & \textbf{Avg.} \\
\midrule
\multirow{3}{*}{\textbf{MLLMs}}
& InternVL3.5-38B & 768 & 28.0 & 32.3 & 32.9 & 42.8 & 34.0 \\
& Qwen3-VL-32B-Instruct & 768 & 32.2 & 32.3 & 38.9 & 52.8 & 39.1 \\
& Gemini 2.5 Pro & 768 & 29.2 & 38.7 & 36.1 & 55.1 & 39.8 \\
\midrule
\multirow{3}{*}{\textbf{RAG}}
& Video-RAG & 64 & 37.6 & 50.0 & 48.3 & 57.8 & 48.4 \\
& LightRAG & 1FPS & 42.4 & 50.6 & 47.0 & 56.8 & 49.2 \\
& HippoRAG & 1FPS & 47.0 & 56.3 & 47.9 & 51.0 & 50.6 \\
\midrule
\multirow{5}{*}{\textbf{\shortstack{Memory\\Agents}}}
& Ego-R1 & 1FPS & 32.0 & 35.3 & 32.9 & 42.8 & 35.8 \\
& HippoMM  & 1FPS & 35.2 & 38.7 & 45.3 & 56.2 & 43.9 \\
& EgoButler & 1FPS & 44.4 & 42.8 & 47.9 & 55.6 & 47.7 \\
& WorldMM & 1FPS & 49.6 & 53.7 & 50.6 & 52.2 & 51.5 \\
& \textbf{EvoMem (Ours)} & 1FPS & \textbf{55.0} & \textbf{61.0} & \textbf{59.1} & \textbf{65.2} & \textbf{60.1} \\
\bottomrule
\end{tabular}
\caption{Performance comparison of EvoMem with baselines across long-video QA benchmarks. Except for direct LVLM baselines, reproduced external-memory methods use Qwen3-VL-32B-Instruct for memory construction, retrieval, and answering.}
\label{tab:main_results}
\end{table*}

EvoMem then evaluates the candidate policy on independent validation episodes to determine whether it substantively improves over the incumbent policy. Let $y_i^{\mathrm{inc}}\in\{0,1\}$ and $y_i^c\in\{0,1\}$ denote the answer correctness of the incumbent policy and the candidate policy on validation sample $i$, respectively. For validation samples where the candidate skill is actually triggered and affects the retrieval process, EvoMem assigns a reward according to the answer-correctness transition:
\begin{equation}
r_i(c)=
\left\{
\begin{array}{ll}
+1, & y_i^{\mathrm{inc}}=0,\ y_i^c=1,\\
-\lambda, & y_i^{\mathrm{inc}}=1,\ y_i^c=0,\\
0, & \mbox{otherwise}.
\end{array}
\right.
\end{equation}
The penalty coefficient $\lambda>1$ introduces an asymmetric penalty, so that performance regressions are penalized more strongly than improvements are rewarded. Functionally, this reward signal plays a role similar to an advantage signal in reinforcement learning: a candidate skill receives positive feedback if it repairs an error made by the incumbent policy, receives a stronger penalty if it breaks a case already solved by the incumbent policy, and receives zero reward if the correctness outcome does not change. The global validation reward of the candidate skill is the cumulative return over the set of validation samples $\mathcal{V}_c$ on which it is actually triggered:
\begin{equation}
R(c)=\sum_{i\in\mathcal{V}_c} r_i(c).
\end{equation}

Finally, EvoMem performs a conservative policy update gated by validation reward. A candidate skill is formally accepted into the active skill bank only when its aggregate validation reward exceeds a positive threshold:
\begin{equation}
\mathcal{B}^{(r+1)}=
\left\{
\begin{array}{ll}
\mathcal{B}^{(r)}\cup\{c\}, & R(c)\ge \tau,\\
\mathcal{B}^{(r)}, & R(c)<\tau.
\end{array}
\right.
\end{equation}
Accepted skills are added to the active skill bank and participate in skill selection and retrieval control for subsequent questions; skills that fail validation are recorded in the rejected skill bank as negative candidate history for future evolution. Through this closed loop of trajectory sampling, policy proposal, advantage evaluation, and conservative update, EvoMem transfers feedback-based policy improvement to the space of retrieval skills, enabling continual and stable self-evolution of long-video memory retrieval policies without updating the parameters of the foundation model.

\section{Experiments}

\subsection{Experimental Setup}
\paragraph{Datasets and Metrics}
We evaluate EvoMem on four long-video question answering benchmarks: EgoLifeQA~\cite{yang2025egolife}, Ego-R1 Bench~\cite{tian2025egor1chainoftoolthoughtultralongegocentric}, LVBench~\cite{Wang_2025_ICCV}, and the long-video subset of Video-MME~\cite{Fu_2025_CVPR}. These benchmarks cover complementary settings, including ultra-long egocentric QA and reasoning, hour-scale temporal understanding, and comprehensive video analysis with multimodal evidence. Following prior work, we report answer accuracy (\%) on each benchmark, and also report the average accuracy over benchmarks with available comparable results.

\paragraph{Baselines}
We compare EvoMem with three groups of representative baselines. The first group consists of general multimodal large language models, including InternVL3.5-38B~\cite{wang2025internvl35advancingopensourcemultimodal}, Qwen3-VL-32B-Instruct~\cite{bai2025qwen3vltechnicalreport}, and Gemini 2.5 Pro~\cite{comanici2025gemini25pushingfrontier}. The second group includes retrieval-augmented methods, including HippoRAG~\cite{gutiérrez2025ragmemorynonparametriccontinual}, LightRAG~\cite{guo2025lightragsimplefastretrievalaugmented}, and Video-RAG~\cite{NEURIPS2025_f593c9c2}, which retrieve external evidence to support long-video reasoning. The third group includes memory-agent methods for long-video reasoning, including Ego-R1~\cite{tian2025egor1chainoftoolthoughtultralongegocentric}, EgoButler~\cite{yang2025egolife}, HippoMM~\cite{lin2026hippomm}, and WorldMM~\cite{yeo2026worldmm}. Missing entries indicate that the corresponding results are unavailable or not reported under the same evaluation setting.

\paragraph{Implementation Details}
We use Qwen3-VL-32B-Instruct as the base model of EvoMem for all test-time components, including video memory construction, skill selection, retrieval decision making, evidence interpretation, and final answer generation. For fair comparisonunder matched context budgets across all methods, we reproduce each external-memory baseline following its official retrieval procedure and implementation settings, with the only modification being the use of Qwen3-VL-32B-Instruct for all LVLM-based components. GPT-5.5 is used only in the skill self-evolution stage to induce candidate skills from successful and failed retrieval trajectories, and is not used as the test-time answering model. For skill self-evolution, we construct an evolution set from public training questions and reserved benchmark questions. For each benchmark, skill evolution and final evaluation use disjoint question sets, and all reported results are obtained on the final-evaluation set.

\subsection{Main Results}
Table~\ref{tab:main_results} presents the main results of EvoMem and representative baselines on four long-video question answering benchmarks. EvoMem achieves the strongest overall performance on all evaluated datasets, reaching accuracies of 55.0\%, 61.0\%, 59.1\%, and 65.2\% on EgoLifeQA, Ego-R1 Bench, LVBench, and Video-MME (L), respectively. Compared with the strongest non-EvoMem baseline on each benchmark, EvoMem brings improvements of 5.4, 4.7, 8.5, and 7.4 percentage points, with an average accuracy of 60.1\%. These results show that the advantage of EvoMem is not limited to a single dataset or video type, but remains consistent across long-video question answering settings with different lengths, viewpoints, and evidence requirements.

\begin{figure}[t]
  \centering
  \includegraphics[width=\linewidth]{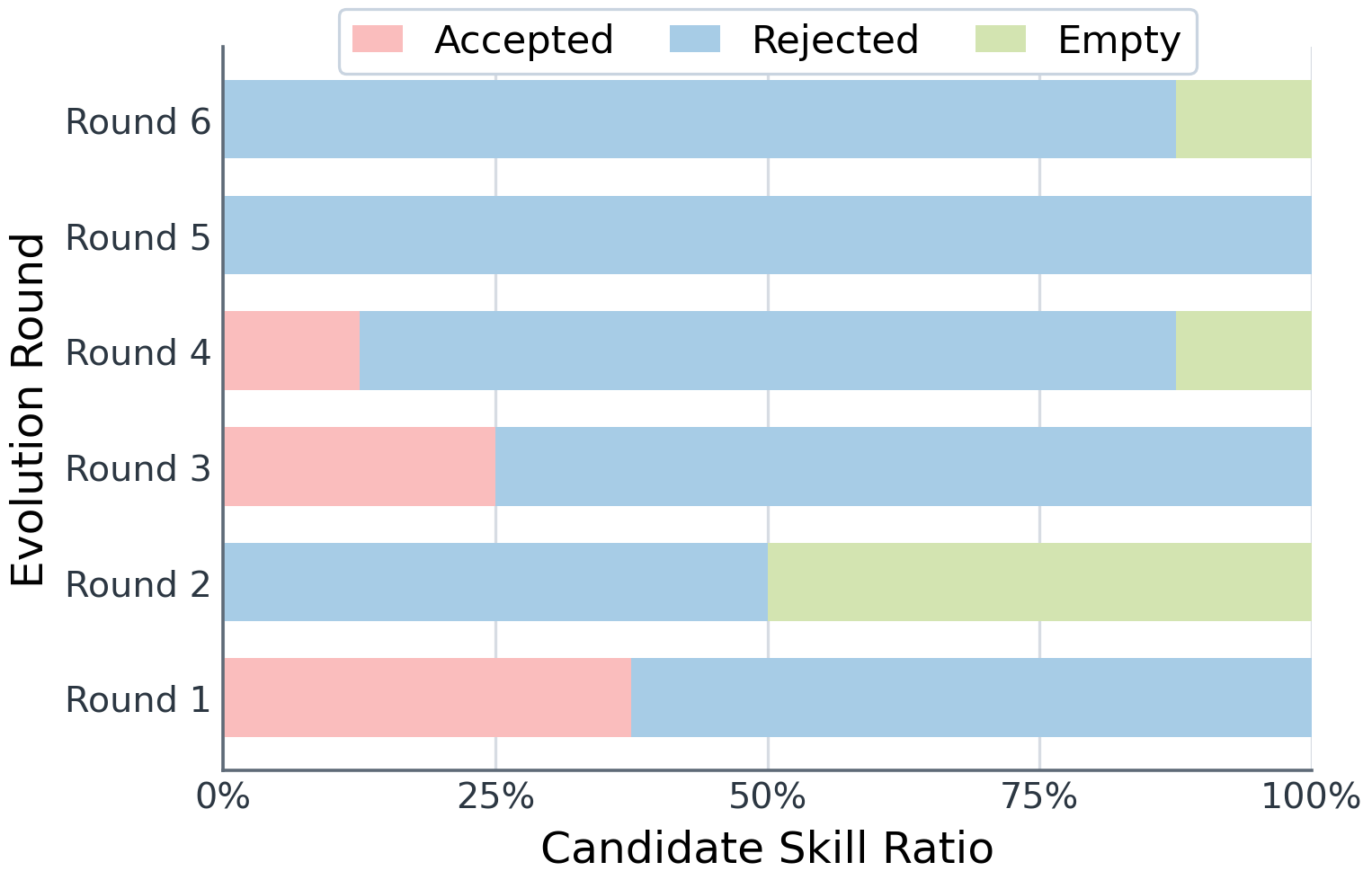}
  \caption{Filtering ratios of candidate skills across LVBench evolution rounds as accepted, rejected, and not generated.}
  \label{fig:3}
\end{figure}

\begin{figure}[t]
  \centering
  \includegraphics[width=\linewidth]{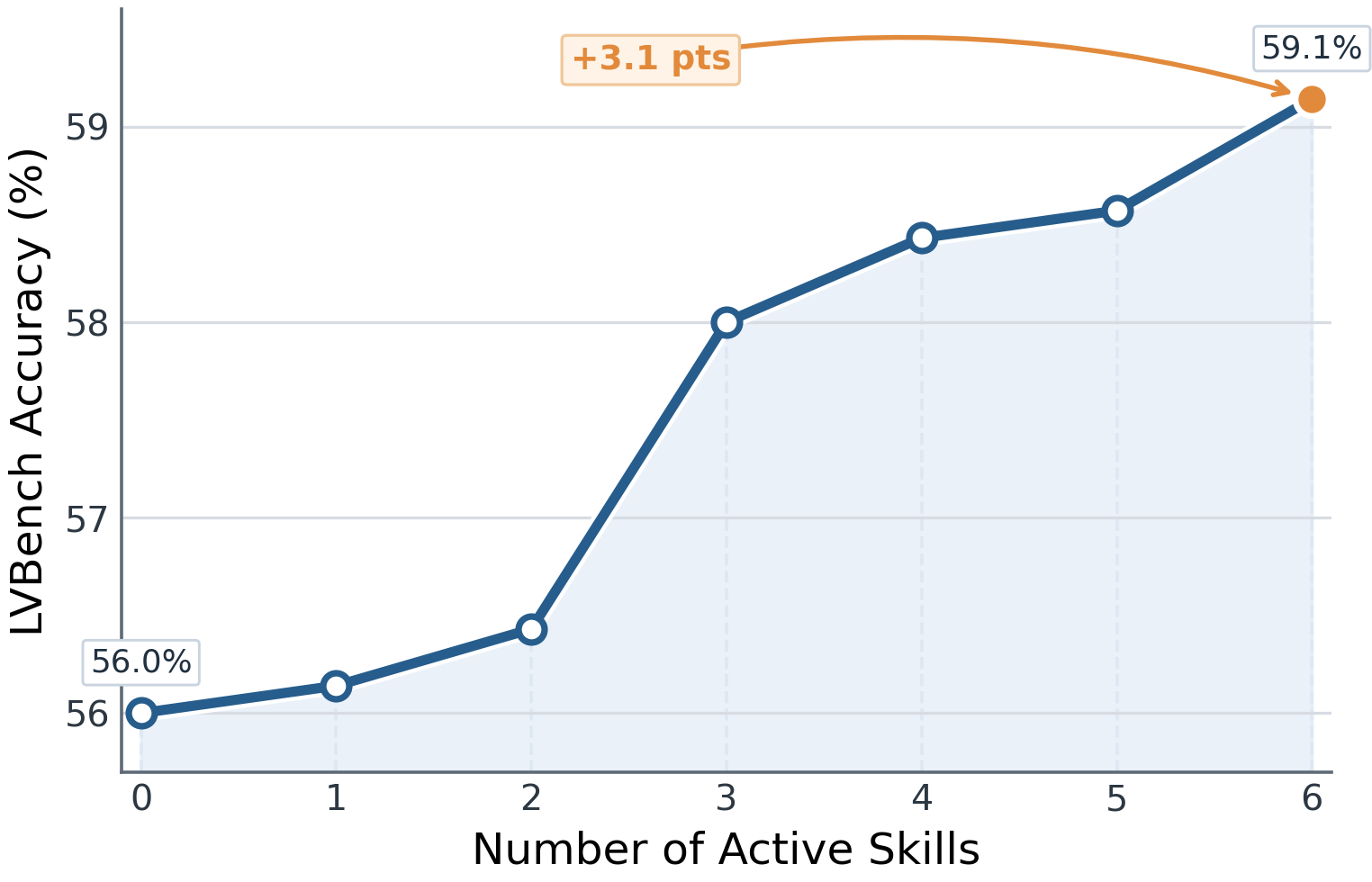}
  \caption{LVBench accuracy as the active skill bank grows.}
  \label{fig:4}
\end{figure}

The results on LVBench further highlight the advantages of EvoMem in time-aware memory access and skill-guided retrieval planning. As an extreme long-video understanding benchmark, LVBench often involves explicit or implicit temporal cues, requiring a model to convert these cues into effective retrieval constraints and use them to locate and verify relevant evidence in long videos. Therefore, LVBench provides a representative setting for evaluating time-aware retrieval. The results show that even with only basic memory operations, EvoMem reaches an accuracy of 56.0\% through time-indexed memories and controllable textual retrieval, interval reading, and visual evidence acquisition, outperforming the strongest baseline in Table~\ref{tab:main_results}, WorldMM, by 5.4 percentage points.Building on this foundation, Fig.~\ref{fig:3} summarizes candidate filtering during skill evolution. After the skill-evolution process, we further evaluate the saved skill-bank snapshots on LVBench to analyze how performance changes as the active skill bank grows. As shown in Fig.~\ref{fig:4}, accuracy increases to 59.1\%, expanding the margin over WorldMM to 8.5 percentage points. The additional gain of 3.1 percentage points over basic memory operations indicates that process-level skill guidance further strengthens the agent's ability to schedule retrieval content, memory operations, and target temporal ranges.

\subsection{Ablation Studies}

\begin{table}[t]
\centering
\small
\setlength{\tabcolsep}{2.4pt}
\begin{tabular}{lccccc}
\toprule
\multirow{2}{*}{\textbf{Method}}
& \textbf{EgoLife} & \textbf{Ego-R1} & \textbf{LV} & \textbf{Video-MME} & \multirow{2}{*}{\textbf{Avg.}} \\
& \textbf{QA} & \textbf{Bench} & \textbf{Bench} & \textbf{(L)} & \\
\midrule
Basic Ops. & 52.8 & 58.7 & 56.0 & 62.8 & 57.6 \\
- Time-Aware & 48.8 & 57.3 & 47.7 & 58.0 & 53.0 \\
+ Human Skill & 54.4 & 59.0 & 57.4 & 63.6 & 58.6 \\
\textbf{+ Evolved Skill} & \textbf{55.0} & \textbf{61.0} & \textbf{59.1} & \textbf{65.2} & \textbf{60.1} \\
\bottomrule
\end{tabular}
\caption{Ablation of EvoMem components.}
\label{tab:ablation}
\end{table}

\begin{table}[t]
\centering
\small
\setlength{\tabcolsep}{4.0pt}
\begin{tabular}{lccc}
\toprule
\multirow{2}{*}{\textbf{Model}} & \multicolumn{3}{c}{\textbf{Ego-R1 Bench}} \\
\cmidrule(lr){2-4}
& No Skill & + Skill & Gain \\
\midrule
EvoMem (Qwen3-VL-8B-Instruct)  & 50.3 & 51.7 & +1.4 \\
EvoMem (Qwen3-VL-32B-Instruct) & 58.7 & 61.0 & +2.3 \\
EvoMem (GPT-5.5)      & 66.3 & 72.3 & +6.0 \\
\bottomrule
\end{tabular}
\caption{Generalization of retrieval skill guidance under different retrieval-model configurations on Ego-R1 Bench.}
\label{tab:skill_transfer_models}
\end{table}

Table~\ref{tab:ablation} evaluates the contributions of time-aware memory access, retrieval skill guidance, and skill self-evolution in EvoMem. Compared with the 53.0\% average accuracy after removing the time-aware mechanism, Basic Ops. with time-aware access reaches 57.6\%, showing that hierarchical time-indexed memories and controllable memory operations provide an effective foundation for long-video evidence acquisition. On top of this, Human Skill improves the average accuracy to 58.6\%, while reward-guided skill self-evolution further raises it to 60.1\%; These results show that skill self-evolution can use validation feedback to identify and retain more effective process-level retrieval strategies, improving the reliability of multi-step evidence acquisition under different evidence requirements. Notably, for egocentric QA, we evolve the retrieval skill bank only on the EgoLifeQA training split and directly apply the frozen skill bank to Ego-R1 Bench. As shown in Table~\ref{tab:ablation}, this skill bank still brings consistent gains on Ego-R1 Bench, providing evidence for the cross-dataset transferability of self-evolved skills.

In addition, Table~\ref{tab:skill_transfer_models} further examines the generalization ability of retrieval skill guidance under different retrieval-model configurations. To avoid the influence of memory construction differences, this experiment uses the same constructed video memories and only changes the test-time model used for retrieval decision making, evidence understanding, and answer generation. The results show that adding skills on top of the no-skill basic operations consistently improves performance across different model configurations. Specifically, Qwen3-VL-8B-Instruct and Qwen3-VL-32B-Instruct obtain gains of 1.4 and 2.3 percentage points, respectively. When using the stronger GPT-5.5 as the test-time retrieval model, the base accuracy reaches 66.3\%, and adding skills further improves it to 72.3\%, yielding a gain of 6.0 percentage points. This trend indicates that the skill-guided retrieval strategy in EvoMem generalizes well across retrieval-model configurations, and that stronger models can better leverage process-level control signals for complex long-video retrieval planning.

\section{Conclusion}
We presented \textbf{EvoMem}, a time-aware long-video memory agent that combines hierarchical time-indexed memories with self-evolving retrieval skills. Experiments show that EvoMem outperforms existing methods, validating time-aware retrieval and skill-guided evidence acquisition for long-video reasoning.

\bibliography{aaai2027}


\end{document}